\documentclass[10pt,letterpaper]{article}
\usepackage[top=0.85in,left=2.75in,footskip=0.75in]{geometry}

\usepackage{amsmath,amssymb}

\usepackage{changepage}

\usepackage[utf8x]{inputenc}

\usepackage{textcomp, marvosym}

\usepackage{cite}

\usepackage{nameref,hyperref}

\usepackage[right]{lineno}

\usepackage{microtype}
\DisableLigatures[f]{encoding = *, family = * }

\usepackage[table]{xcolor}

\usepackage{array}

\newcolumntype{+}{!{\vrule width 2pt}}

\newlength\savedwidth

\usepackage{setspace} 
\raggedright
\usepackage[aboveskip=1pt,labelfont=bf,labelsep=period,justification=raggedright,singlelinecheck=off]{caption}

\makeatletter
\renewcommand{\@biblabel}[1]{\quad#1.}
\makeatother

\usepackage{lastpage,fancyhdr,graphicx}
\usepackage{epstopdf}
\fancyheadoffset[L]{2.25in}
\fancyfootoffset[L]{2.25in}
\usepackage{makecell}
\usepackage{soul}
\usepackage{gensymb}

\begin{document}
\vspace*{0.2in}

\begin{flushleft}
{\Large
\textbf\newline{Deep convolutional neural networks for segmenting 3D \textit{in vivo} multiphoton images of vasculature in Alzheimer disease mouse models} 
}
\newline
\\
Mohammad Haft-Javaherian\textsuperscript{1},
Linjing Fang\textsuperscript{1},
Victorine Muse\textsuperscript{1},
Chris B. Schaffer\textsuperscript{1},
Nozomi Nishimura\textsuperscript{1},
Mert R. Sabuncu\textsuperscript{1,2*},
\\
\bigskip
\textbf{1} Nancy E. and Peter C. Meinig School of Biomedical Engineering, Cornell University, Ithaca, NY, USA
\\
\textbf{2} School of Electrical and Computer Engineering, Cornell University, Ithaca, NY, USA
\\
\bigskip

%
%





* msabuncu@cornell.edu

\end{flushleft}
\section*{Abstract}
The health and function of tissue rely on its vasculature network to provide reliable blood perfusion. Volumetric imaging approaches, such as multiphoton microscopy, are able to generate detailed 3D images of blood vessels that could contribute to our understanding of the role of vascular structure in normal physiology and in disease mechanisms. The segmentation of vessels, a core image analysis problem, is a bottleneck that has prevented the systematic comparison of 3D vascular architecture across experimental populations. We explored the use of convolutional neural networks to segment 3D vessels within volumetric \textit{in vivo} images acquired by multiphoton microscopy. We evaluated different network architectures and machine learning techniques in the context of this segmentation problem. We show that our optimized convolutional neural network architecture with a customized loss function, which we call \textit{DeepVess}, yielded a segmentation accuracy that was better than state-of-the-art methods, while also being orders of magnitude faster than the manual annotation. To explore the effects of aging and Alzheimer's disease on capillaries, we applied \textit{DeepVess} to 3D images of cortical blood vessels in young and old mouse models of Alzheimer's disease and wild type littermates. We found little difference in the distribution of capillary diameter or tortuosity between these groups, but did note a decrease in the number of longer capillary segments ($>75\mu m$) in aged animals as compared to young, in both wild type and Alzheimer's disease mouse models.



\section*{Introduction} 
The performance of organs and tissues depend critically on the delivery of nutrients and removal of metabolic products by the vasculature. Blood flow deficits due to disease related factors or aging often leads to functional impairment~\cite{sonntag2007regulation}. In particular, the brain has essentially no energy reserve and relies on the vasculature to provide uninterrupted blood perfusion~\cite{hossmann1994viability}. 

Multiple image modalities can be used to study vascular structure and dynamics, each offering tradeoffs between the smallest vessels that can be resolved and the volume of tissue that can be imaged. Recent work with several modalities, including photoacoustic microscopy~\cite{lin2017high}, optical coherence tomography~\cite{erdener2017spatio}, and multiphoton microscopy (MPM)~\cite{blinder2013cortical}, enable individual capillaries to be resolved in 3D over volumes approaching 1 $mm^3$ in living animals. The analysis of such images is one of the most critical and time-consuming tasks of this research, especially when it has to be done manually. 

For example, in our own work we investigated the mechanisms leading to reduced brain blood flow in mouse models of Alzheimer’s disease (AD), which required extracting topology from capillary networks each with $\sim 1,000$ vessels from dozens of animals. The manual tracing of these networks required $\sim 40\times$ the time required to acquire the images, greatly slowing research progress~\cite{Cruz2019Neutrophil}. The labor involved in such tasks limits our ability to investigate the vital link between capillary function and many different diseases. Many studies have shown anatomical and physiological differences in microvasculature associated both with age and AD, such as changes in composition of large vessel walls' smooth muscles~\cite{kalaria1996cerebral}, increased collagen VI in microvascular basement membranes and their thickening in AD~\cite{farkas2001cerebral}, and age-associated reduction of microvascular plasticity and the ability of the vessels to respond appropriately to changes in metabolic demand~\cite{riddle2003microvascular}.

In this paper, we consider the segmentation of vessels, a core image analysis problem that has received considerable attention~\cite{kirbas2004review,lesage2009review}. As in other segmentation and computer vision problems, in recent years deep neural networks (DNNs) have offered state-of-the-art performance~\cite{litjens2017survey}. DNN approaches often rely on formulating the problem as supervised classification (or regression), where a neural network model is trained on some (manually) labeled data. For a survey on deep learning in medical image analysis, see a recent review by Litjens et al.~\cite{litjens2017survey}. 

Here, we explore the use of a convolutional neural network (CNN) to segment 3D vessels within volumetric \textit{in vivo} MPM images. 
\textit{In vivo} MPM imaging of blood vessels has the advantage that it captures the size and shape of vessels without introducing artifacts from postmortem tissue processing. However, blood flow generates features which must be accommodated in the vessel segmentation. 
We conduct a thorough study of different network architectures and machine learning techniques in the context of this segmentation problem. We apply the final model, which we call \textit{DeepVess}, on image stacks of cortical blood vessels in mouse models of AD and wild type (WT) littermates. Our experimental results show that \textit{DeepVess} yields segmentation accuracy that is better than current state-of-the-art, while being orders of magnitude faster than the manual annotation (20-30 hours manual work vs. 10 minutes computation time). 
The segmentation method developed in this work provides robust and efficient analysis which enabled us to quantify and compare capillary diameters and other vascular parameters from \textit{in vivo} cortex images across multiple animals, with varying age as well as across WT mice and AD models. 

\section*{Related work}
Blood vessel segmentation is one of the most common and time-consuming tasks in biomedical image analysis. This problem can either be approached in 2D or 3D, depending on the specifics of the application and analytic technique. The most established blood vessel segmentation methods are developed for 2D retinography~\cite{fraz2012blood} and 3D CT/MRI~\cite{lesage2009review}. 

Among segmentation methods, region-based methods are well-known for their simplicity and low computational cost~\cite{tsai2009correlations}. For example, Yi et al.~\cite{yi2003locally} developed a 3D region growing vessel segmentation method based on local cube tracking. 
In related work, Mille et al.~\cite{mille2009deformable} used a 3D parametric deformable model based on the explicit representation of a vessel tree to generate centerlines. 
In recent years, these traditional segmentation methods have become less popular and are considered to be limited in comparison to deep learning methods, because they require handcrafted filters, features, or logical rules and often yield lower accuracy.

Today, in problems that are closely related to ours, various deep learning techniques dominate state-of-the-art. For instance, in a recent Kaggle challenge for diabetic retinopathy detection within color fundus images, deep learning was used by most of the 661 participant teams, including the top four teams. Interestingly, those top four methods surpassed the average human accuracy. Subsequently, Gulshan et al. \cite{gulshan2016development} adopted the Google Inception V3 network~\cite{szegedy2016rethinking} for this task and reached the accuracy of seven ophthalmologists combined. For retinal blood vessel segmentation, Wu et al.~\cite{wu2016deep} used a CNN-based approach to extract the entire connected vessel tree. Fu et al.~\cite{fu2016retinal} proposed to add a conditional random fields (CRF) to post-process the CNN segmentation output. They further improved their method by replacing the CRF with a recurrent neural network (RNN), which allows them to train the complete network in an end-to-end fashion~\cite{fu2016deepvessel}. Further, Maninis et al. ~\cite{maninis2016deep} addressed retinal vessel and optic disc segmentation problems using one CNN network and could surpass the human expert.

There are 3D capillary image datasets in mice~\cite{tsai2009correlations} and human~\cite{lorthois2014tortuosity} that were segmented using traditional segmentation methods and have illustrated the scientific value of such information, but few such datasets are available.

To the best of our knowledge, there are only two studies that used deep learning for our problem: vascular image analysis of multi-photon microscopy (MPM) images. 
The first one is by Teikari et al.~\cite{teikari2016deep} who proposed a hybrid 2D-3D CNN architecture to produce state-of-the-art vessel segmentation results in 3D microscopy images. The main limitation of their method was the use of 2D convolutions and 2D conditional random fields (CRF)s, which restrict the full exploitation of the information along the third dimension. The second study was conducted by Bates et al.~\cite{bates2017extracting}, where the authors applied a convolutional long short-term memory RNN to extract 3D vascular centerlines of endothelial cells. Their approach was based on the U-net architecture~\cite{ronneberger2015u}, which is a well-known fully convolutional network~\cite{long2015fully} widely used for biomedical image segmentation. Bates and colleagues achieved state-of-the-art results in terms of centerline extraction; nevertheless, they reported that certain vessels in the images were combined in the automatic segmentation.
Finally, we consider the 3D U-Net \cite{cciccek20163d}, which is the volumetric version of the U-net architecture \cite{ronneberger2015u} and is regarded by many as state-of-the-art for microscopy image segmentation problems.

\section*{Data and methods} 
The proposed vasculature segmentation method for 3D \textit{in vivo} MPM images, \textit{DeepVess}, consists of (i) pre-processing to remove \textit{in vivo} physiological motion artifacts due to respiration and heartbeat, (ii) applying a 3D CNN for binary segmentation of the vessel tree, and (iii) post-processing to remove artifacts such as network discontinuities and holes. 
\subsection*{Data}
\subsubsection*{Animals} 
All animal procedures were approved by the Cornell University Institutional Animal Care and Use Committee and were performed under the guidance of the Cornell Center for Animal Resources and Education. We used double transgenic mice (B6.Cg-Tg (APPswe, PSEN1dE9) 85Dbo/J, referred to as APP/PS1 mice) that express two human proteins associated with early onset AD, a chimeric mouse/ human amyloid precursor protein (Mo/HuAPP695swe) and a mutant human presenilin1 (PS1-dE9), which is a standard model of AD and typically develops amyloid-beta plaque deposition around 6 months of age~\cite{jankowsky2003mutant}. Littermate WT mice (C57BL/6) served as controls. Animals were of both sexes and ranged in age from 18 to 31 weeks for young mice and from 50 to 64 weeks for the old mice (6 WT and 6 AD at each age, for a total of 24 mice). 
\subsubsection*{\textit{In vivo} imaging of cortical vasculature} 
We use a locally-designed multiphoton microscope~\cite{denk1990two} for \textit{in vivo} imaging of the brain vasculature. Glass-covered craniotomies were prepared over parietal cortex, as described previously~\cite{holtmaat2009long,shih2012two,Cruz2019Neutrophil}.
For cranial window implantation and imaging, mice were anesthetized with 3\% isoflurane and then maintained on 1.5\% isoflurane in 100\% oxygen. Mice were injected with 0.05 mg/100g of mouse weight glycopyrrolate (Baxter Inc.) or 0.005 mg/100g atropine (intramuscular 54925-063-10, Med-Pharmex Inc.). At time of surgery as well as 1 and 2 days after mice received 0.025 mg/100g dexamethasone (subcutaneous 07-808-8194, Phoenix Pharm Inc.), and 0.5 mg/100g ketoprofen (intramuscular, Zoetis Inc.). Bupivacaine (0.1 ml, 0.125\%, Hospira Inc.) was subcutaneously injected at the incision site. Animals were injected with 1 ml/100g mouse 5\% (w/v) glucose in normal saline subcutaneously every hour during imaging and surgery. Body temperature was maintained at 37\degree C with a feedback-controlled heating blanket (40-90-8D DC, FHC). Mice were euthanized with pentobarbital overdose after their last imaging session.

We waited at least three weeks after the surgery before imaging to give time for the mild surgically-induced inflammation to subside. Windows typically remained clear for as long as 20 weeks. This technique allows us to map the architecture of the vasculature throughout the top $500\ \mu m$ of the cortex. Briefly, the blood plasma of an anesthetized mouse was labeled with an intravenous injection of Texas Red labeled dextran (70 KDa, Life Technologies). The two-photon excited fluorescence intensity was recorded while the position of the focus of a femtosecond laser pulse train was scanned throughout the brain, providing a three-dimensional image of the vasculature~\cite{denk1990two}. Imaging was done using 800-nm or 830-nm, 75-fs pulses from a Ti:Sapphire laser oscillator (MIRA HP, pumped by a Verdi-V18, or Vision S, Coherent). Lasers were scanned by galvonometric scanners and focused into the sample using a 1.0 NA, 20X water-immersion objective lens (Carl Zeiss, Inc.). Image stacks were acquired with 645/45 nm (center wavelength/bandwidth) bandpass filters. The ScanImage software package~\cite{pologruto2003scanimage} was used to control the whole system. Image stacks were taken with a range of magnifications resulting in lateral voxel sizes from 0.45 to 1.71 $\mu m$/pixel, but always 1 $\mu m$ in the axial direction. 
\subsubsection*{Expert annotation}
We implemented a protocol to facilitate the manual 3D segmentation task using ImageJ, an open-source image processing software package~\cite{abramoff2004image} (supplementary material). Two people, one expert and one less experienced, each manually segmented a motion artifact corrected (see below), $256\times256\times200$ voxels ($292 \times 292 \times 200 \ \mu m^3$) image from an AD mouse, independently, which took about 20 and 30 hours, respectively. The second annotator was trained by the expert and then had several months of practice prior to performing this task.
These data were used to estimate inter-human segmentation variation. We treated the expert labels as the ``gold standard'' segmentation and used the second annotator's labels to compare variability in manual segmentation. All other comparisons were made with respect to the gold standard segmentation as the ground truth. This dataset was divided into independent (i.e., non-overlapping) training, validation, and testing sub-parts (50\%-25\%-25\%), all spanning the entire depth of the stack. The training and validation datasets were used in the optimization of CNN architectures, while the test dataset was kept unused until the end of our architecture design optimization process and used for the final unbiased evaluation. We repeated this process 4 times, by varying the test data and thus effectively conducting 4-fold cross-validation. We note that architecture optimization was only done in the first fold. Additionally, six independent 3D images (different mice and different voxel size) acquired by Cruz Hern{\'a}ndez et al. \cite{Cruz2019Neutrophil} were labeled by an expert to examine the generalization of \textit{DeepVess}. The detailed properties of these images are in Table~\ref{tab.SecondImageInfo}. With this paper, we also have made all images and expert annotations publicly available at: https://doi.org/10.7298/X4FJ2F1D
\subsection*{Preprocessing}
Motion artifacts caused by physiological movements are one of the major challenges for 3D segmentation of \textit{in vivo} MPM images. Furthermore, global linear transformation models cannot compensate for the local nonuniform motion artifacts, for example, due to a breath occurring part way though the raster scanning for an MPM image. In this study, we adopted the non-rigid non-parametric diffeomorphic demons image registration tool implemented based on the work of Thirion~\cite{thirion1998image} and Vercauteren et al.~\cite{vercauteren2009diffeomorphic}. Our approach is to register each slice to the previous slice, starting from the first slice as the fixed reference. The diffeomorphic demons algorithm aims to match the intensity values between the reference image and deformed image, where cost is computed as the mean squared error. The smoothness prior on the deformation field is implemented via an efficient Gaussian smoothing of gradient fields, and invertibility is ensured via concatenation of small deformations. This kernel is effectively encouraging the deformation field to be smooth, thus regularizing the ill-posed non-linear registration problem. Based on our experiments, a Gaussian kernel with the standard deviation of 1.3 was chosen for the regularization of the registration algorithm. Next, in our pre-processing steps, the 1-99\% range of the image intensities in the input image patch were linearly mapped between 0 and 1, and the extreme 1\% of voxels were clipped at 0 and 1. This step, we found, helps with generalizing the model to work well with images taken from other MPM platforms by adapting normalization parameters to the acquisition systems and image statistics utilizing most of the intensity rang. To facilitate comparison between different datasets, image volumes were resampled to have 1 $\mu m^3$ voxel for comparisons.
\subsection*{Convolutional neural network architectures}
Our aim in this work is to design a system that takes an input stack of images (in 3D) and produces a segmentation of vessels as a binary volume of the same size. 
For this task, as we elaborate below, we explored different CNN architectures using validation performance as our guiding metric. 
Our baseline CNN architecture starts with a 3D input image patch (tile), which has $33\times33\times5$ voxels (in x, y, and z directions). 
The first convolution layer uses a $7\times7\times5$ voxel kernel with 32 features to capture 3D structural information within the neighborhood of the targeted voxel. The output of this layer, 32 nodes of $27\times27\times1$ voxel images, enter a max pooling layer with a $2\times2$ kernel and $2\times2$ strides. Another convolution layer with $5\times5\times1$ kernel and 64 features, followed by a similar max pooling layer are then applied before the application of the fully connected dense layer with 1024 hidden nodes and dropout~\cite{srivastava2014dropout} with a probability value of 50\%. The output is a two-node layer, which represents the probability that the pixel at the center of the input patch belongs to tissue vs. vessel. The CNN takes an input 3D patch and produces a segmentation label for the central voxel. All the convolution layers have a bias term and rectified linear unit (ReLU) as the element-wise nonlinear activation function. Starting from this baseline CNN architecture, we optimized the network architecture hyperparameters with a greedy algorithm.

Different kernel sizes for the 3D convolution layers were explored in our experiments. Note that each choice in the architecture parameters (including the kernel size) corresponded to a different input patch size. As the validation results summarized in Table~\ref{tab.Results-FOV} indicate, the best performing baseline architecture had an input patch size of $33\times33\times7$. 
Based on this result we chose an input patch size of $33\times33\times7$ as the optimal field of view (FOV) for segmentation. 
We then explored the effect of the number of convolutional and max pooling layers. As summarized in Table~\ref{tab.Results-Architecture}, the best architecture had three 3D convolution layers with a $3\times3\times3$ voxel kernel, a max pooling layer, followed by two convolution layers with a $3\times3$ voxel kernel, and a max pooling layer.
The output of the last max pooling layer is reshaped to a fully-connected layer followed by a 1024-node fully-connected layer and the last fully-connected layer, which is reshaped to the output patch size. Note that there is no difference in spatial resolution (i.e., voxel dimensions) between the input and output patches.

\begin{figure}[!h]
	\centering
	\includegraphics[width=\linewidth]{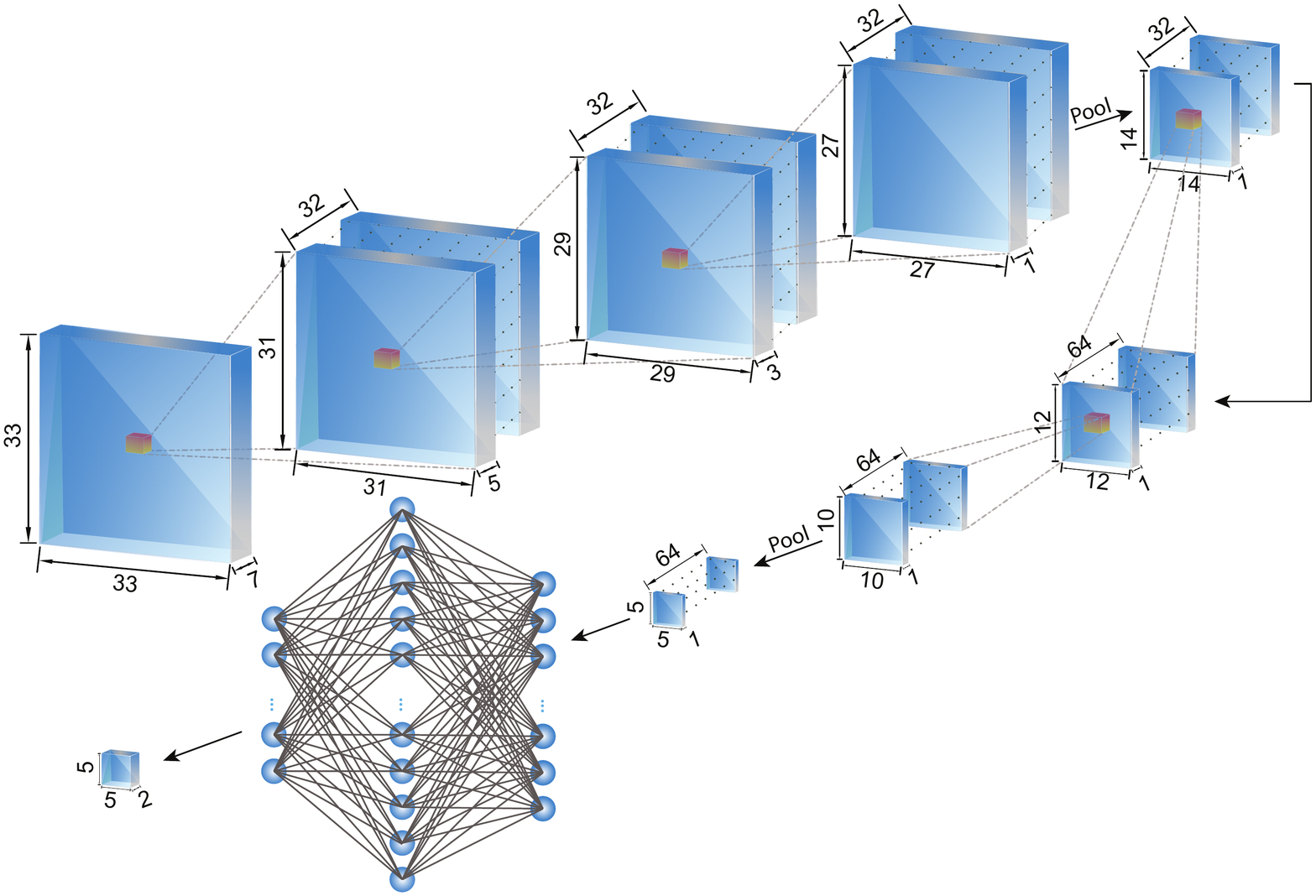}
	\caption{The optimal 3D CNN architecture. The field of view (FOV), i.e. the input patch size, is $33\times33\times7$ voxels and the output is the segmentation of the $5\times5\times1$ patch (region of interest, ROI) at the center of the patch. The convolution kernels are $3\times3\times3$ voxels for all the layers and ReLU is used as the element-wise nonlinear activation function. The first three convolution layers have 32 channels and are followed by pooling. The second three convolution layers have 64 channels. The output of convolution layers is $5\times5\times1$ voxels with 64 channels, which is fed to a fully connected neural network with a 1024-node hidden layer. The final result has $5\times5\times1$ voxels with two channels representing the probability of the foreground and background label associations.}
	\label{fig1:CNN architecture scheme}
\end{figure}

Finally, we investigated the performance for different output patch sizes, ranging from 1 voxel to $5\times5\times5$ voxels and found that performance was improved further when the output is the segmentation of the central $5\times5\times1$ patch and not just a single voxel. 
A larger output area has the advantage of accounting for the structural relationship between adjacent voxels in their segmentation. 
The optimal CNN architecture scheme is shown in Fig~\ref{fig1:CNN architecture scheme}. 

\subsection*{Performance metrics}
There are different performance metrics to compare agreement between an automated segmentation method and a ``ground truth'' (GT) human annotation. 
In the context of binary segmentation, the foreground ($F$) will be the positive class, and the negative class will correspond to the background ($B$). 
Therefore, true positive (TP) can be measured as the total number of voxels where both the automatic and human segmentation labels are foreground.
True Negative (TN), False Positive (FP) and False Negative (FN) can be defined in a similar fashion.

Based on these, we can compute \textit{sensitivity} and \textit{specificity}. For example, sensitivity is the percentage of GT foreground voxels that are labeled by the automatic segmentation (ASeg) correctly. 
Mathematically, we have: 
\begin{eqnarray} 
sensitivity &= P(y=F|GT=F) = \frac{TP}{TP + FN} \\
specificity &= P(y=B|GT=B) = \frac{TN}{TN + FP} 
\end{eqnarray}
The \textit{Dice coefficient} (DC), \textit{Jaccard index} (JI), and \textit{modified Hausdorff distance} (MHD) are another set of commonly used segmentation performance metrics. JI is defined as the ratio between the number of voxels labeled as foreground by both GT and ASeg, to the total number of voxels that are called foreground by either GT and ASeg. DC is very similar to JI, except it values TP twice as much as FP and FN. JI and DC are useful metrics when the number of the foreground voxels is much less than background and the detection accuracy of the foreground voxels is more important compared to background voxel detection, which is the case for 3D imaging of vasculature. 
\begin{eqnarray} 
JI = P\left(y=F \cap GT=F \ | \ y=F \cup GT=F\right) = \frac{TP}{TP + FP + FN} \\
DC = \frac{2\times JI}{1+JI} = \frac{2\times TP}{2\times TP + FP + FN}
\end{eqnarray}
On the other hand, MHD~\cite{dubuisson1994modified} quantifies accuracy in terms of distances between boundaries, which might be appropriate when considering tubular structures. For each boundary point in image \textbf{A} ($a\in\mathcal{A}$), the closest Euclidean distance ($d(a,b)=||a-b||_2$) to any boundary point inside image \textbf{B} ($b\in\mathcal{B}$) is first calculated, $d(a,\mathcal{B}) = min_{b\in\mathcal{B}}\ ||a-b||_2$). This is then averaged over all boundary points in $\mathcal{A}$: $\frac{1}{N_a}\sum_{a\in\mathcal{A}}^{}d(a,\mathcal{B})$ ~\cite{huttenlocher1993comparing}. MHD is then defined as:
\begin{eqnarray} 
MHD = max\left[\frac{1}{N_a}\sum_{a\in\mathcal{A}}^{}d(a,\mathcal{B}), \ \frac{1}{N_b}\sum_{b\in\mathcal{B}}^{}d(b,\mathcal{A})\right] \\
d(a,\mathcal{B}) = min_{b\in\mathcal{B}}\ ||a-b||_2 
\end{eqnarray}	
Note that in the segmentation setting, \textbf{A} and \textbf{B} can represent the foreground boundaries in the automatic and GT segmentations, respectively.
Finally, we can compute the MHD on centerlines instead of boundaries, a metric we call MHD-CL.
\subsection*{Training and implementation details}
In training our segmentation algorithms, we used a customized cross-entropy loss function designed for our highly unbalanced datasets (where foreground voxels comprise only a small fraction of the volume), measured over all voxels but TN ($i\in \{TP,FP,FN\}$), defined as: 
\begin{eqnarray} 
Loss = \sum_{i\in \{TP,FP,FN\}} -\left[y_i\ \log(p_i)+(1-y_i)\log(1-p_i)\right] \label{eq:loss}
\end{eqnarray}
$y_i$ is the GT label and $p_i$ is the model's output as the probability of the target voxel \textit{i} belonging to the foreground. Note that in Eq. (\ref{eq:loss}), true negative voxels have no contribution, effectively reducing the influence of the dominant background. We trained our model using \textit{Adam} stochastic optimization~\cite{kingma2014adam} with a learning rate of $10^{-4}$ for 100 epochs during architecture exploration and a learning rate of $10^{-6}$ for 30,000 epochs during the fine tuning of model parameters for the proposed architecture with mini-batch size of 1000 samples (based on GPU memory constraints and results of our experiments with smaller mini-batch size, which did not improve the optimization results). The fine tuning took one month on one NVIDIA TITAN X GPU. We implemented our models in Python using Tensorflow\textsuperscript{\texttrademark}~\cite{abadi2016tensorflow}. 
\subsection*{Post-processing}
CNN segmentation results contain some segmentation artifact such as holes inside the vessels, rough boundaries, or isolated small objects. In order to remove these artifacts, the holes within the vessels were filled. This was followed by application of a 3D mean filter with a $3\times3\times3$ voxel kernel and the removal of small foreground objects, e.g. smaller than 100 voxels. This result was used to compare to the gold standard.
\subsection*{Analysis of vasculature centerlines}
To characterize the cortical vasculature of the experimental animals, we identified capillary segments by calculating centerlines from the segmented image data. 
Our centerline extraction method includes dilation and thinning operations, in addition to some centerline artifact removal steps. 
The binary segmentation image was first thinned using the algorithm developed by Lee et al.~\cite{lee1994building}. The result was then dilated using a spherical kernel with a radius of 5-voxels to improve the vessel connectivity, which was followed by mean filtering with a $3\times3\times3$ voxel kernel and removing holes from each cross section. Next, a thinning step was applied again to obtain the new centerline result. 
The original segmented image was dilated using a spherical kernel with a radius of 1-voxel to act as the mask for the centerlines with the goal of improving the centerline connectivity. The following rules were applied to the resulting centerlines repeatedly until no further changes could be done. A vessel is a segment between two bifurcations. 
\begin{enumerate}
	\item Remove any vessels with one end not connected to the network (i.e., dead end) and with length smaller than 11 voxels.
	\item Remove single voxels connected to a junction.
	\item Remove single voxels with no connections.
	\item Remove vessel loops with length of one or two voxels. 
\end{enumerate}
Finally, the centerline network representation (i.e. nodes, edges, and their properties) was extracted. (The centerline extraction was applied on both manual and automated segmentations.)

\section*{Results} 	
\begin{figure}[!h]
	\centering
	\includegraphics[width=\linewidth]{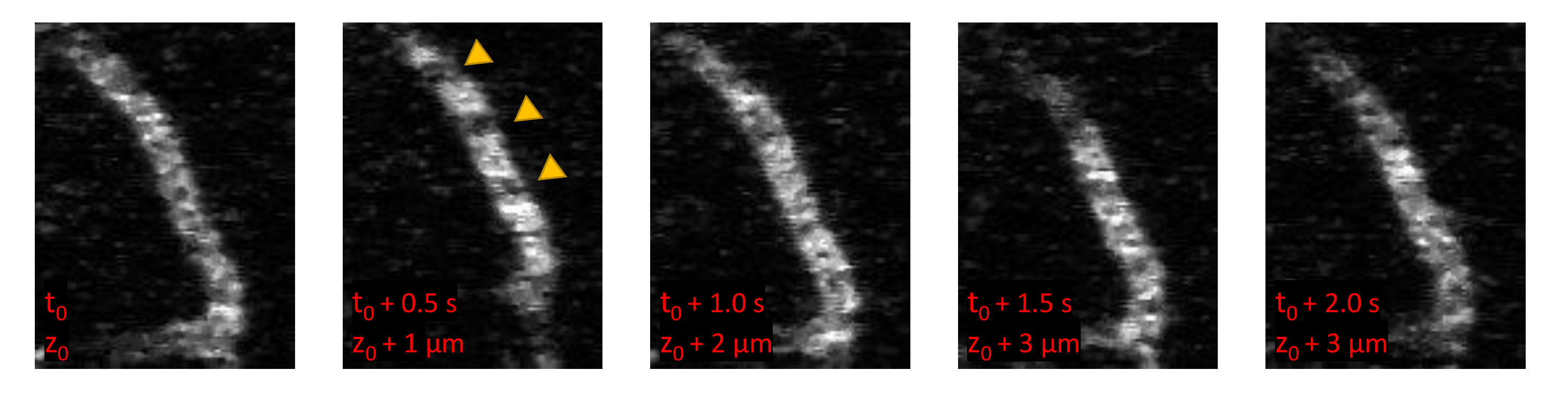}
	\caption{\textit{In vivo} MPM images of a capillary. Because MPM images are acquire by raster scanning, images at different depths (z) are acquired with a time lag (t). Unlabeled red blood cells moving through the lumen cause dark spots and streaks and result in variable patterns within a single vessel.}
	\label{fig2:RBC_moving}
\end{figure}

We conducted a systematic evaluation of several network architecture parameters in order to optimize segmentation accuracy of images of mouse cortex vasculature from MPM. 
Features of \textit{in vivo} MPM images include motion artifacts due to respiration and heart beat. Because vessels are visualized by an injection of dye that labels the blood plasma, unlabeled red blood cells appear as dark spots and streaks moving through the vessel lumen (arrows in Fig~\ref{fig2:RBC_moving}). Images are acquired by raster scanning through the tissue and each vessel is captures in several images. The imaging speed has a significant influence on these features and in \textit{in vivo} experiments, imaging is often relatively slow, such that these features become prominent~\cite{jones2018vivo}.
We emphasize that our exploration was based on performance on the validation dataset and the final results presented reflect the model accuracy on an independent test dataset.
The detailed performance results for some of the tested architectures are reported in Tables~\ref{tab.Results-FOV} and \ref{tab.Results-Architecture}. 
The optimal architecture, \textit{DeepVess}, was trained on the training data until the model accuracy stopped improving and no overfitting was observed (30,000 epochs). Fig~\ref{figS1.JI_vs_batch} shows the JI learning curve over 30,000 epochs, for training, validation, and test datasets. The constant gap between JI of the training and validation datasets, which represent generalization error, confirms that we are not strongly overfitting.

Furthermore, we implemented two state-of-the-art methods \cite{teikari2016deep,cciccek20163d}, and an improved version of the method of Teikari et al.~\cite{teikari2016deep}, where we changed the 2D convolutional kernels into 3D kernels and inserted a fully connected neural network layer at the end, based on the suggestion in the discussion of their paper. Table~\ref{tab.Results} summarizes the comparison between the performance of our optimal architecture based on the 4-fold cross-validation results, with and without the post-processing step, comparing to two state-of-the-art methods and a second human annotator to provide a measure of the inter-human variability. These results, as well as Fig~\ref{figS1.JI_vs_batch} demonstrate that \textit{DeepVess} outperforms the state-of-the-art methods \cite{teikari2016deep,cciccek20163d} in terms of sensitivity, Dice index, Jaccard index, and boundary modified Hausdorff distance; and approaches human performance in terms of Dice and Jaccard. The proposed method does not outperform the benchmarks in specificity, indicating a slightly higher rate of false positive voxels. Yet we note that the relatively lower specificity is still very high (97\%).

\begin{table}[!h]
	\begin{adjustwidth}{-2.25in}{0in}
	\caption{The comparison of our proposed CNN architecture (\textit{DeepVess}), manual annotation by a trained person, and two state-of-the-art methods \cite{teikari2016deep, cciccek20163d} to the gold standard of the expert human annotation based on the 4-fold cross-validation results. \textit{DeepVess} surpassed both human annotator and two state-of-the-art methods in terms of sensitivity as well as Dice index, Jaccard index, and boundary modified Hausdorff distance, which are the three metrics that are widely used in segmentation.}
	\begin{center}
		\begin{tabular}{|l|c|c|c|c|c|c|}
			\hline
			\textit{} &\textit{ Sensitivity} & \textit{ Specificity} &  \textbf{\textit{Dice}} & \textbf{\textit{Jaccard}} & \textbf{\textit{MHD}} \\ \hline
			\makecell[cl]{Second human annotator} & 81.07\% & 98.70\% & 82.35\% & 70.40\% & 1.50 \\ \hline
			\makecell[cl]{Original Teikari et al.~\cite{teikari2016deep} } & 62.44\% & \textbf{98.65\%} & 69.69\% & 55.06\% & 3.20 \\ \hline
			\makecell[cl]{Çiçek et al. \cite{cciccek20163d}} & 70.01\% & 98.21\% & 72.69\% & 59.41\% & 3.55 \\ \hline
			\makecell[cl]{Improved \cite{teikari2016deep} in this study} & 69.55\% & 98.39\% & 74.03\% & 59.96\% & 3.16 \\ \hline
			\textit{DeepVess} & 89.91\% & 97.00\% & 81.62\% & 69.13\% & 2.26 \\ \hline
			\makecell[cl]{\textit{DeepVess} with post-processing} & \textbf{89.95\%} & 97.00\% & \textbf{81.63\%} & \textbf{69.15\%} & \textbf{2.25} \\ \hline
		\end{tabular}
	\end{center}
	\label{tab.Results}
\end{adjustwidth}
\end{table}

\begin{figure}[!h]
	\centering
	\includegraphics[width=0.65\linewidth]{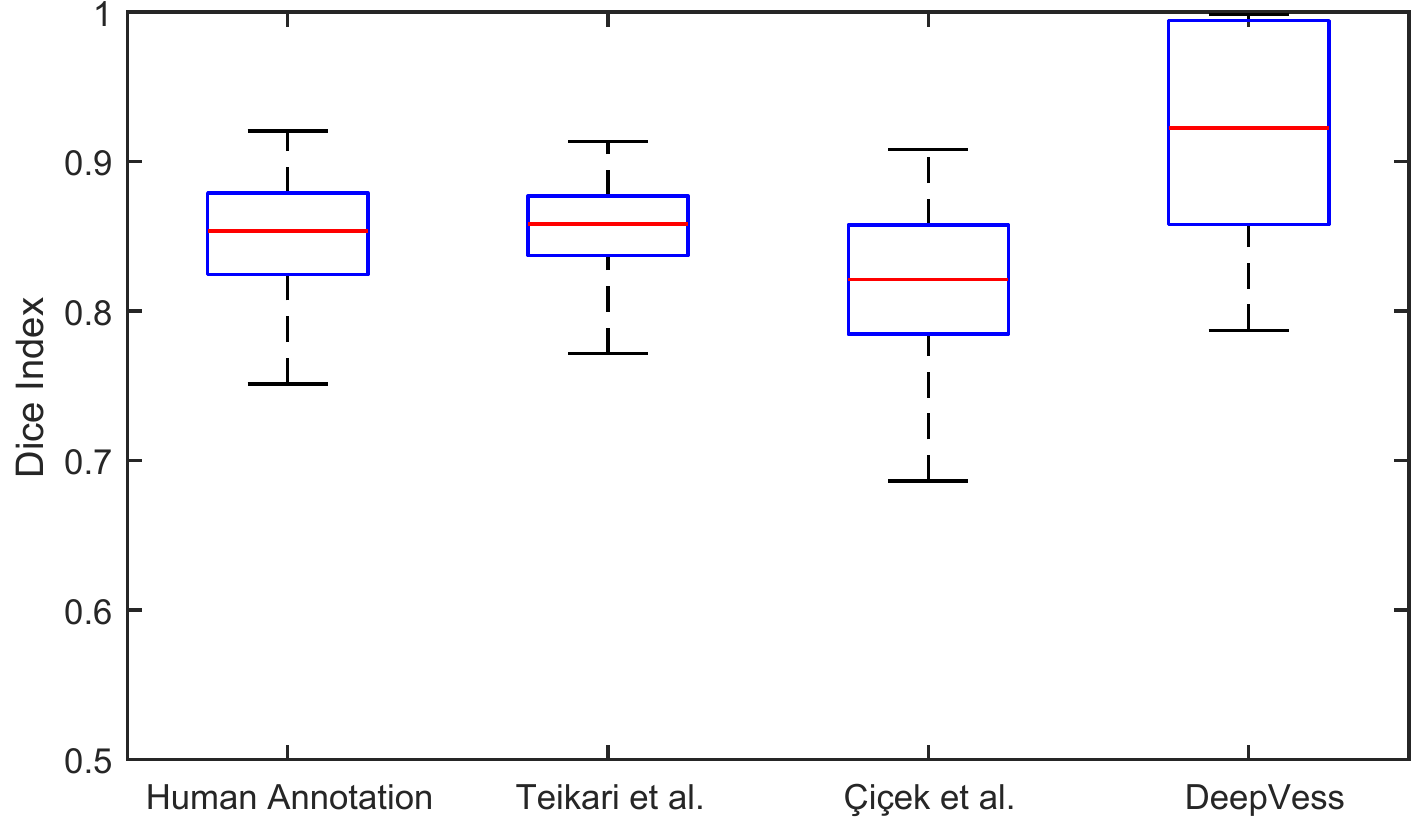}
	\caption{Slice-wise Dice index of \textit{DeepVess} vs. manual annotation by a trained person and the state-of-the-art methods \cite{teikari2016deep,cciccek20163d} compared to the gold standard of the expert human annotation. The central red mark is the median, and the top and bottom of the box is the third and first quartiles, respectively. The whiskers indicates the range of data. \textit{DeepVess} has higher median value in comparison to the Teikari et al.~\cite{teikari2016deep}, {\c{C}}i{\c{c}}ek et al.~\cite{cciccek20163d}, and the human annotator (Wilcoxon signed-rank test, $p=2.98e-23$, $p=2.59e-32$, and $p=2.8e-28$, respectively).}
	\label{fig3.Dice_Boxplot}
\end{figure}

In MPM, the variation in the signal to noise as a function of imaging depth leads to changes in image quality between image slices. 
The performance of a segmentation method should therefore be assessed by analyzing slices separately. 
Fig~\ref{fig3.Dice_Boxplot} illustrates the boxplot of slice-wise Dice index values from the x-y planes within the 3D MPM image dataset. 
\textit{DeepVess} had a higher Dice index values in comparison to the Teikari et al. and the trained annotator's results. 
However, there was more variation compared to the other two results, which implies the possibility and need for further improvements. 

\begin{figure}[!h]
	\centering
	\includegraphics[width=\linewidth]{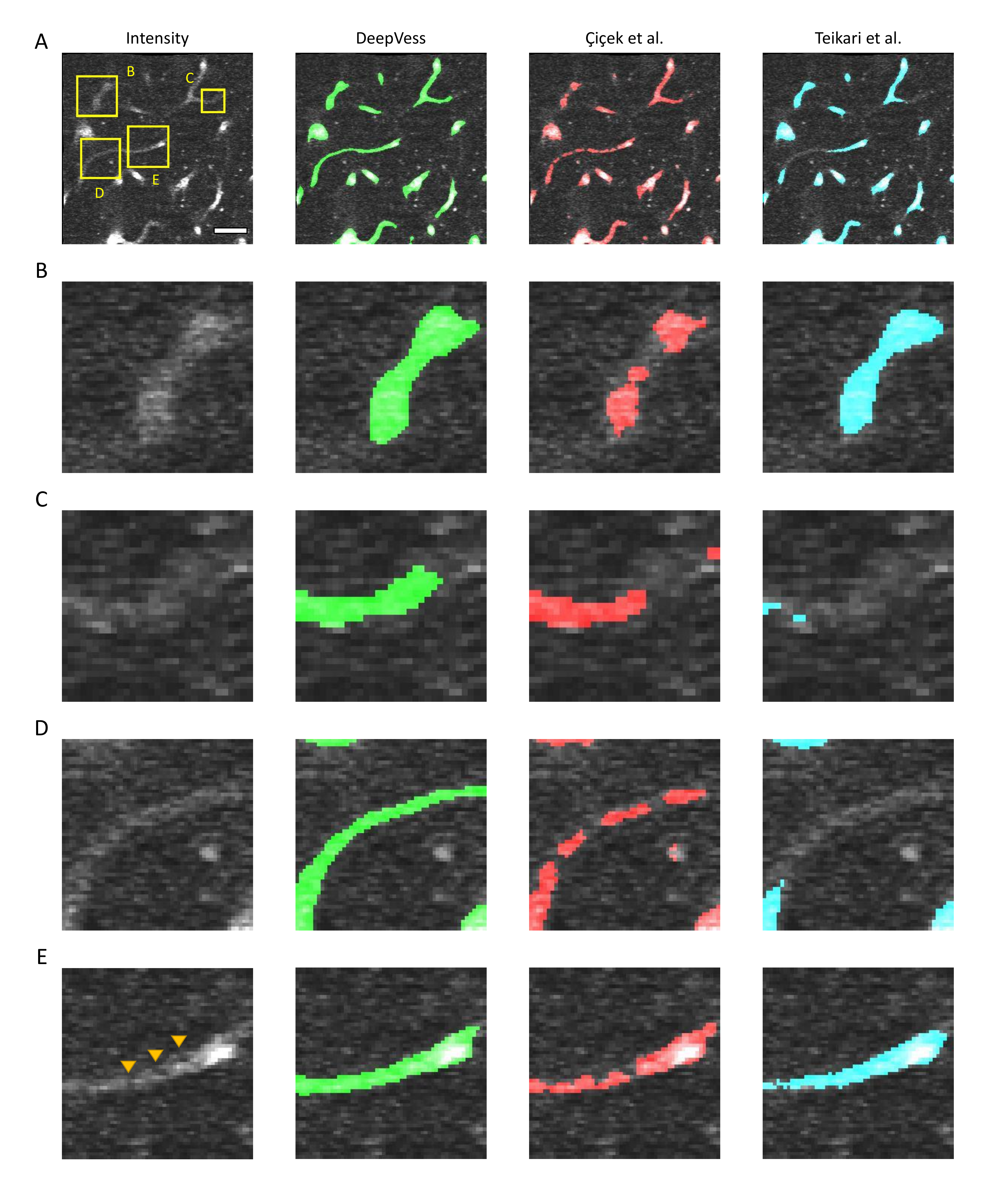}
	\caption{Comparison of \textit{DeepVess} and the state-of-the-art methods \cite{teikari2016deep,cciccek20163d} in a 3D image cross-section obtained from an independent dataset (Table~\ref{tab.SecondImageResults} \#1) not used during the training. (A) An image frame with intensity in gray and overlay of segmentation from each method. (B-E) magnified view of four cases within A. The three models overlaid on the complete 3D image is made available online in Supplemental Materials. Scale bar is $50\mu m$.}
	\label{fig4.model_comparison}
\end{figure}

The generalization of the model was studied by testing an independent dataset annotated by our expert consisting of 6 separate 3D MPM images acquired from 1 AD and 5 WT mice (Table~\ref{tab.SecondImageInfo}) and the results are summarizes in Table~\ref{tab.SecondImageResults}.
\textit{DeepVess} outperforms both the state-of-the-art methods \cite{teikari2016deep,cciccek20163d} on the second dataset in terms of sensitivity, Dice index, Jaccard index, and boundary MHD. Similar to the test dataset results, specificity was slightly lower. These results illustrate the generalization of our model on new MPM images with different image quality and captured from different mouse models and with different voxels sizes.
Fig~\ref{fig4.model_comparison}.A illustrates the image intensity and three models overlaid on the image for a cross-section extracted from a 3D image from the independent dataset (Table~\ref{tab.SecondImageResults} \#1). Fig~\ref{fig4.model_comparison}.B-E are magnified version of three cases within Fig~\ref{fig4.model_comparison}.A . The main sources of failure in the vessel segmentations of 3D \textit{in vivo} MPM images are low SNR at deeper cross-sections (Fig~\ref{fig4.model_comparison}.C) and unlabeled, moving red blood cells in the vessel lumen, which cause dark spots and streaks (Fig~\ref{fig4.model_comparison}.B and D). The patchy segmentations due to unlabeled red blood cells result in unconnected and isolated vasculature centerlines and network. 
The \textit{DeepVess} architecture has fully connected layers and thus might be exploiting some spatially varying properties of the signal (as in the variation of contrast as a function of depth) that a fully convolutional architecture such as U-Net might not be able to exploit. 	
Elsewhere, in the absence of such  difficulties, all three models segment the vessels largely accurately. 

We next examined the quality of the vessel centerlines derived from the different segmentations. Using the centerline modified Hausdorff distance (CL MHD) as a centerline extraction accuracy metric, \textit{DeepVess} (CL MHD [\textit{DeepVess}] = 3.03) is substantially better than the state-of-the-art methods (CL MHD [Teikari et al.] = 3.72, CL MHD [Çiçek et al.] = 6.13). But there is still room for improvement in terms of automatic centerline extraction as neither automatic methods yielded scores as good as the trained human annotator (CL MHD [human annotator] = 2.73). In order to test the accuracy of geometrical measurements, the vessel diameter, a sensitive metric, was selected. We measured the diameter of 100 vessels manually by averaging ten 2D measurements per vessel to compare with the \textit{DeepVess's} results (Fig~\ref{figS2.manual diameter}). We observed that there is no significant difference between manually measured diameters and \textit{DeepVess's} results (paired t-test, $n=100$, $p=0.34$).

\section*{Discussion} 
The segmentation of 3D vasculature images is a laborious task that slows down the progress of biomedical research and constrains the use of imaging in clinical practice. 
There has been significant research into tackling this problem via image analysis methods that reduce or eliminate human involvement. 
In this work, we presented a CNN approach, which surpasses the state-of-the-art vessel segmentation methods~\cite{teikari2016deep,cciccek20163d} as well as a trained human annotator. 
The proposed algorithm, \textit{DeepVess}, segments 3D \textit{in vivo} vascular MPM images with more than ten million voxels in ten minutes on a single NVIDIA TITAN X GPU, a task that takes 30 hours for a trained human annotator to complete manually. 

\begin{figure}[!h]
	\centering
	\includegraphics[width=\linewidth]{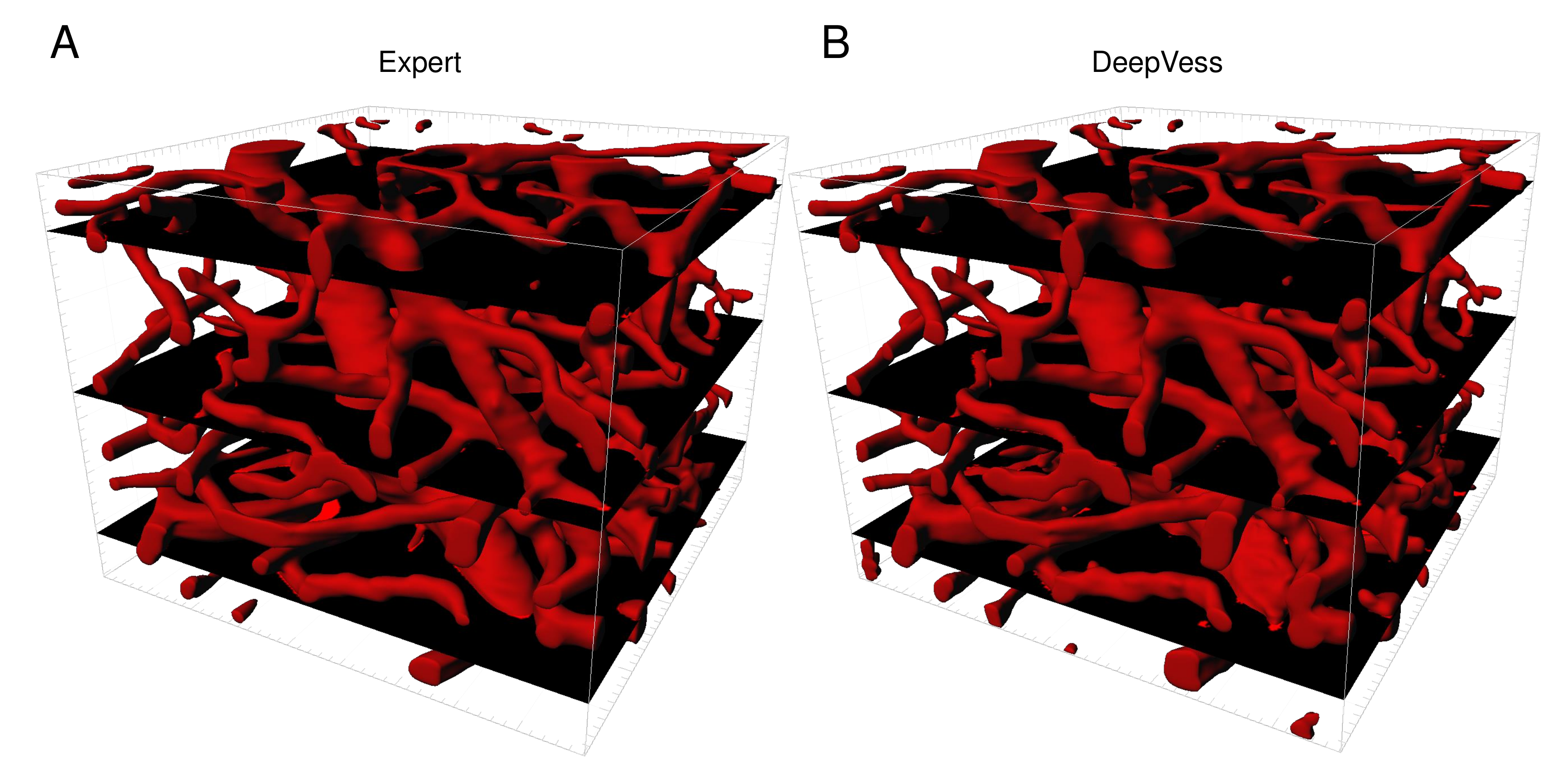}
	\caption{3D rendering of (A) the expert's manual and (B) \textit{DeepVess} segmentation results. The top, middle, and bottom black plains correspond to the high, medium, and low quality examples, respectively, which are analyzed further in the Discussion (Fig~\ref{fig6.Results}). Each volume is $256\times256\times200$ voxels ($292 \times 292 \times 200 \ \mu m^3$).}.
	\label{fig5.3D_render}
\end{figure}

In order to characterize the performance of \textit{DeepVess}, we compared the automated segmentation to an expert manual segmentation (Fig~\ref{fig5.3D_render}). Here, we visualized three slices with different qualities of segmentation results. 
The 3D rendering of the mouse brain vasculature shown in Fig~\ref{fig5.3D_render} indicates the location of these top, middle, and bottom slices representing typical high, medium, and low segmentation quality, respectively. Top layers are very similar, and differences are visible at the bottom layers, which have low SNR.

\begin{figure}[!h]
	\centering
	\includegraphics[width=\linewidth]{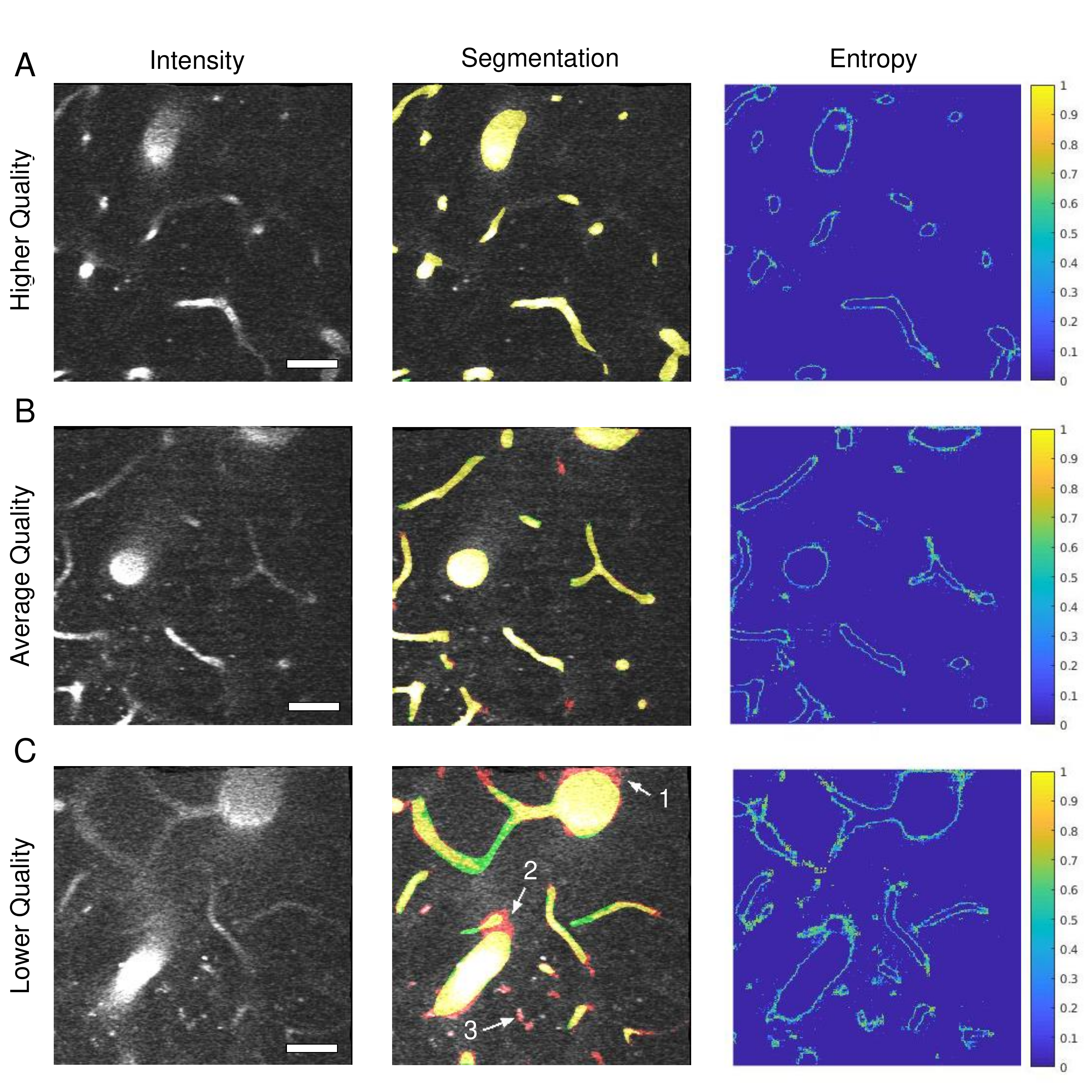}
	\caption{Comparison of \textit{DeepVess} and the gold standard human expert segmentation results in image planes as shown in Fig~\ref{fig5.3D_render}. Imaging is generally higher quality at planes closer to the sample surface. (Left column) Image intensity shown with gray scale after motion artifact removal. The dark spots within the vessels are red blood cells that do not take up the injected dye. (Middle column) Comparison between \textit{DeepVess} (red) and the expert (green) segmentation results overlaid on images. Yellow shows agreement between the two segmentations. (Right column) Shannon entropy, which is a metric of \textit{DeepVess} segmentation uncertainty computed with 50\% dropout at test-time~\cite{gal2016dropout}. The boundaries of vessels with high entropy values, shown in warmer colors, demonstrate the uncertainty of \textit{DeepVess} results at those locations. Scale bar is $50\mu m$.}
	\label{fig6.Results}
\end{figure}

We used 50\% dropout during test-time~\cite{gal2016dropout} and computed Shannon's entropy for the segmentation prediction at each voxel to quantify the uncertainty in the automated segmentation. 
Higher entropy represents higher segmentation uncertainty at a particular voxel. 
The entropy results together with the comparison between \textit{DeepVess} and the expert segmentations for those three planes are illustrated in Fig~\ref{fig6.Results}. 
The left column contains the intensity gray-scale images of these examples. The segmentation results of the \textit{DeepVess} and the expert are superimposed on the original gray-scale image with red (\textit{DeepVess}) and green (the expert), as shown in the middle column. Yellow represents agreement between \textit{DeepVess} and the expert. 
The right column shows the entropy of each example estimated via test time dropout. 
We observe that, in general, \textit{DeepVess} has higher uncertainty at the boundaries of vessels. 
The disagreement with ground truth is also mostly concentrated at the boundaries. 
Images from deeper within the brain tissue that often have lower image contrast and higher noise levels due to the nature of MPM, suffer from more segmentation errors. 
These images can often be challenging even for expert humans. 
Arrows in Fig~\ref{fig6.Results}.C highlight examples of these difficulties. 
The error example 1 illustrates the case where the expert ignored bright pixels around the vessel lumen based on their knowledge of the underlying physiology and experience with MPM images of brain that postulate a rounded lumen instead of a jittery and rough lumen, despite a very strong signal. 
The error example 2 illustrates a low intensity vessel junction that was judged to be an artifact by humans based on experience or information from other image planes.
The error example 3 illustrates the case where a small vessel does not exhibit a strong signal and it is not connected to another major vessel. 

\textit{DeepVess} implements pre- and post-processing tools to deal with \textit{in vivo} MPM images that suffer from different motion artifacts.
\textit{DeepVess} is freely available at \textit{https://github.com/mhaft/DeepVess} and can be used immediately by researchers who use MPM for vasculature imaging. 
Also, our model can be fine-tuned further by adjusting the intensity normalization step to utilize a different part of the intensity range and training samples for other 3D vasiform structures or other imaging modalities.
Similar to many machine learning solutions, \textit{DeepVess's} performance depends on specific image features and the performance will degrade in cases where the tissues are labeled differently (e.g. vessel walls are labeled instead of blood serum) or the images intensities are concentrated in a small portion of the intensity range.

Although \textit{in vivo} measurements present unique challenges to image segmentation, such as the red blood cell motion, in our case, we have shown that \textit{DeepVess} successfully handles these challenges. Postmortem techniques all change the vessel diameters in the tissue processing. Hence, we believe that \textit{in vivo} imaging is the best strategy to quantify vessel diameters. While features such as topology and length might not be affected by postmortem processing, \textit{in vivo} imaging with MPM is important for capillary diameter measurements. Two-photon microscopy has been used to validate histology in many studies (\cite{so1998two,so2000two,tsai2003all,zoumi2004imaging,blinder2013cortical,tsai2009correlations}) and comparisons with other labeling techniques are quite common.

While \textit{DeepVess} offers very high accuracy in the problem we consider, there is room for further improvement and validation, in particular in the application to other vasiform structures and modalities. 
For example, other types of (e.g., non-convolutional) architectures such as long short-term memory (LSTM) can be examined for this problem. 
Likewise, a combined approach that treats segmentation and centerline extraction methods together, such as the method proposed by Bates et al.~\cite{bates2017extracting} in a single complete end-to-end learning framework might achieve higher centerline accuracy levels.

\section*{Application to Alzheimer’s mouse models}
\subsection*{Capillary alteration caused by aging and Alzheimer's disease}				
\textit{In vivo} imaging with multiphoton microscopy of capillary beds is free of distortions in vessel structure caused by postmortem tissue processing that can result in artifacts such as altered diameters~\cite{blinder2013cortical}. However, the images often suffer from poor signal to noise and motion artifacts. An additional challenge is that unlabeled, moving red blood cells in the vessel lumen cause dark spots and streaks that move over time. Disease models are often especially challenging because inflammation and tissue damage can further degrade imaging conditions.  

Strong correlations between vascular health, brain blood flow and AD suggest that mapping the microvascular network is critical to the understanding of cognitive health in aging~\cite{iadecola2004neurovascular}. To explore this question, we imaged the cortical vascular networks in young and old mouse models of AD (young AD and old AD) and their young and old WT littermates (young WT and old WT). Imaged volumes ranged from $230\times 230$ to $600\times 600 \ \mu m^2$ in x-y and 130 to 459 $\mu m$ in the z direction. We imaged 6 animals per group, with at least 3000 capillary segments analyzed for each group.

\begin{figure}[!h]
	\centering
	\includegraphics[width=\linewidth]{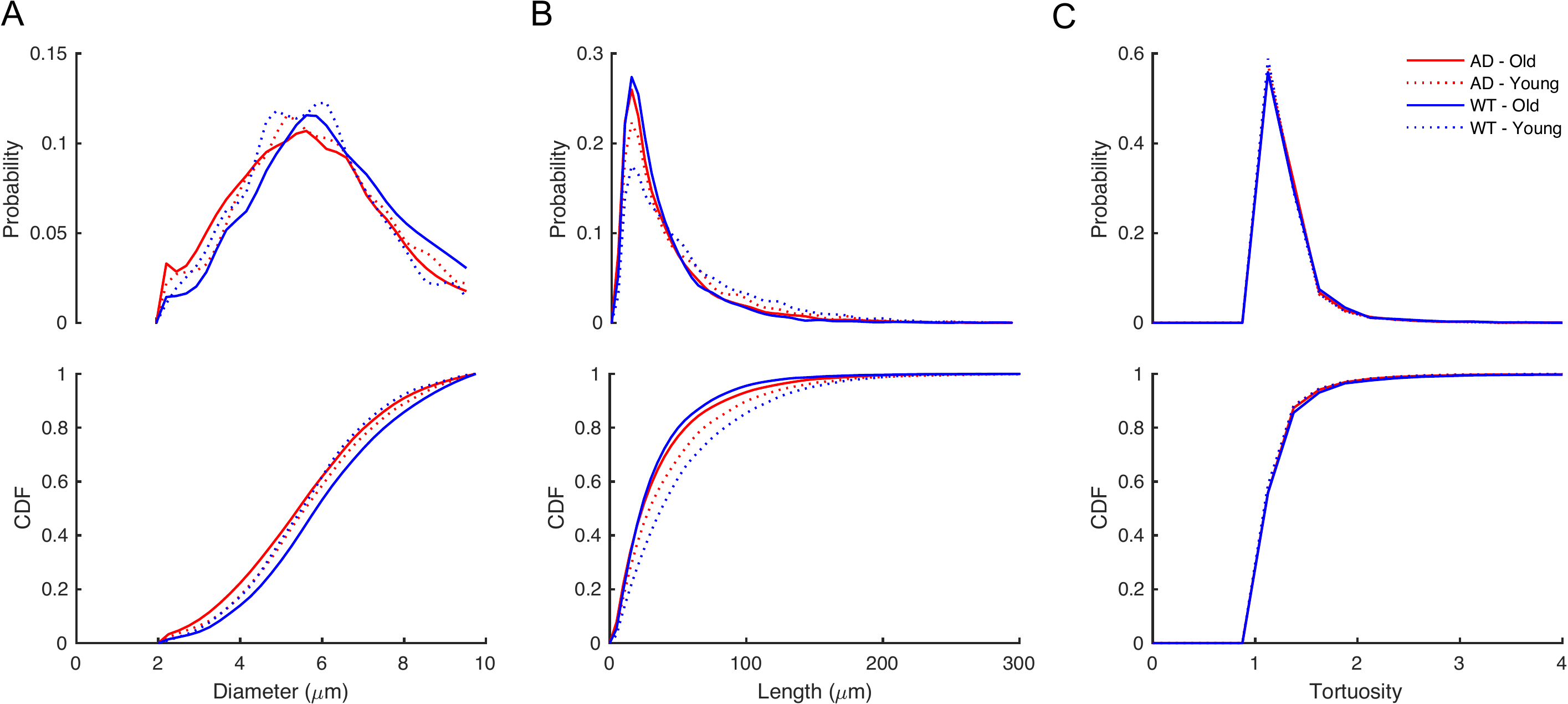}
	\caption{Comparison of capillaries between young and old mice with WT and AD genotype (6 mice in each group). The relative probability and cumulative distribution function (CDF) of the (A) diameters, (B) length, and (C) tortuosity based on all capillaries aggregated within each of the four groups. We compared these metrics between the groups using Kruskal-Wallis test followed by Bonferroni multiple comparison correction~\cite{milliken2009analysis} (Table~\ref{tab.ks_resutl}).} 
	\label{fig7.ADvsWT}
\end{figure}

The resulting 3D stacks of images were preprocessed, segmented with \textit{DeepVess}, and post-processed as discussed in the previous sections. Centerlines were extracted and individual vessel segments were identified. To analyze capillaries while excluding arterioles and venules, only vessel segments less than $10~\mu m$ in diameter were included \cite{verant2007direct,gutierrez2016effect,Cruz2019Neutrophil}. 
For the vascular parameters of segment length, diameter, and tortuosity considered here, previous work has shown that AD mouse models have increased tortuosity in cortical penetrating arterioles as compared to WT mice \cite{dorr2012amyloid,lai2015venular}. Our analysis of capillaries excluded these vessels.
Three metrics were selected to characterize the vascular network. For each capillary segment, we calculated the diameter averaged along the length (Fig~\ref{fig7.ADvsWT}.A), the length (Fig~\ref{fig7.ADvsWT}.B), and the tortuosity, defined as the length divided by the Euclidean distance between the two ends (Fig~\ref{fig7.ADvsWT}.C). The distributions of capillary diameter, length, and tortuosity varied little between young and old mice or between WT and AD genotype (Table~\ref{tab.ks_resutl}). There were subtle shifts ($\sim 0.25~\mu m$) in the diameter distribution between groups, but no clear differences across old/young or WT/AD and the differences in means were small compared to the standard deviation (6-27\% of SD). 
However, we observed a decrease in the number of longer length ($>75\mu m$) capillaries in older animals as compared to young in both WT and AD mice shown by a rightward shift in the cumulative distribution function curve (Fig~\ref{fig7.ADvsWT}.B and Table~\ref{tab.ks_resutl}). 

\begin{table}[!h]
	\begin{adjustwidth}{-2.25in}{0in}
		\begin{center}
			\caption{Comparison between metrics distributions between different groups using Kruskal-Wallis test followed by Bonferroni multiple comparison correction. $\Delta \mu$ is the difference between the mean values of the two tested groups.}
			\begin{tabular}{|l|l|l|l|l|l|l|}
				\hline
				\multicolumn{ 1}{|c|}{} & \multicolumn{ 2}{c|}{\textbf{Diameter ($\mu m$)}} & \multicolumn{ 2}{c|}{\textbf{Length ($\mu m$)}} & \multicolumn{ 2}{c|}{\textbf{Tortuosity}} \\ \cline{ 2- 7}
				\multicolumn{ 1}{|l|}{} & \textbf{$\Delta \mu$} & \textbf{P-value} & \textbf{$\Delta \mu$} & \textbf{P-value} & \textbf{$\Delta \mu$} & \textbf{P-value} \\ \hline
				AD-Old vs. 
				AD-Young & 0.206 & 2.61E-7 & 7.908 & 7.5E-22 & 0.016 & 0.798 \\ \hline
				AD-Old vs. 
				WT-Old & 0.475 & 2.93E-27 & 2.787 & 0.055 & 0.019 & 0.645 \\ \hline
				AD-Old vs. 
				WT-Young & 0.095 & 1.20E-5 & 16.16 & 6.9E-67 & 0.018 & 0.321 \\ \hline
				AD-Young vs.
				WT-Old & 0.269 & 6.39E-9 & 10.69 & 1.14E-27 & 0.035 & 0.027 \\ \hline
				AD-Young vs. 
				WT-Young & 0.110 & 0.012 & 8.252 & 9.12E-17 & 1.50E-3 & 1.000 \\ \hline
				WT-Old vs. 
				WT-Young & 0.379 & 1.1E-14 & 18.95 & 2.1E-63 & 0.037 & 0.036 \\ \hline
			\end{tabular}
			\label{tab.ks_resutl}
		\end{center}
	\end{adjustwidth}
\end{table}

\subsection*{Aging and Alzheimer's disease have little effect on capillary characteristics}
Using a large database of vessel segments measured in three dimensions, we surprisingly found only very small differences between groups that were dwarfed by the variance in capillary diameter or tortuosity between young and old animals or between WT and AD mouse models. 
The automation provided by \textit{DeepVess} enabled the evaluation across a very large number of vessels in a large group size.
The strong agreement between the measurements based on DeepVess and the manual measurements by Cruz Hern{\'a}ndez et al.~\cite{Cruz2019Neutrophil}, confirms that the proposed pipeline yields unbiased and accurate metrics to analyze capillary segments. 
There was a decrease in the number of long capillary segments in the aged animals compared to young in both the WT and AD groups. 
Note that the reported metrics only represent the parietal region of cortex and that regional variability can affect our results. These finding may not generalize across all ages and mouse models of AD and could be different in other regions of the brain. Sonntag et al.~\cite{sonntag2007regulation} argue that changes in vasculature due to aging might be non-linear and multi-phasic. For instance, two studies showed that the capillary density increases during adulthood and then declines in more advanced age~\cite{wilkinson1981quantitative,hunziker1979aging}.
Several previous studies have characterized the average diameters of cortical capillaries in mice, as summarized in Table~\ref{tab.diameter_in_literature}, show high variability in results, suggesting that methodological variations make comparison between studies difficult. Other studies that compared AD models and WT also found negligible or no difference in capillary diameters. Heinzer et al. compared a different mouse model (APP23) using MRA and found no difference between WT and AD mice~\cite{heinzer2004hierarchical}. The same group also compared the effects of ``VEGF overexpression'' model and WT using SR$\mu$CT and also found little difference~\cite{heinzer2008novel}. 

There are a wide range of imaging approaches used in these various studies and data from both live animal and postmortem analysis is included.
It is possible that some of these differences emerge when tissues are processed rather than measured \textit{in vivo} as was done here. Studies based on sectioned tissue sample the 3D vascular architecture differently so it is difficult to make direct comparisons between datasets. Measures of capillaries depend on the definition of capillaries. Here it was based on a threshold diameter of $10 \mu m$, which could explain some of the variability in the literature. 
Not surprisingly given the differences in approach and sample preparation, there is significant disagreement between reported average diameters. Some differences may, however, reflect differences in vasculature across strains and ages of animals.

Therefore, the proposed fully automated objective segmentation of 3D \textit{in vivo} images of the vasculature can be used to reduce the variability due to sample preparation and imaging/analysis approach, allowing such strain and age differences to be elucidated clearly.

\begin{table}[!h]
	\begin{adjustwidth}{-2.25in}{0in}
	\caption{Comparison of measured mouse capillary diameters from different studies.}
	\begin{tabular}{|l|l|l|l|l|l|l|} 
		\hline
		\textbf{Study} & \textbf{Background} & \textbf{Trans gene} & \textbf{Phenotype} & \textbf{\makecell[cl]{Age \\ (week)}} & \textbf{Imaging Modality} & \textbf{\makecell[cl]{Vessel \\ Diameter}} \\ \hline
		\textbf{This study} & C57/BL6 & - & WT & 18-31 & \textit{in vivo }2PEF & $5.81 \pm 1.62 \ \mu m$ \\ \hline
		\textbf{This study} & C57/BL6 & - & WT & 50-64 & \textit{in vivo }2PEF & $6.19 \pm 1.76 \ \mu m$ \\ \hline
		\textbf{This study} & C57/BL6 & APP/PS1 & AD & 18-31 & \textit{in vivo }2PEF & $5.92 \pm 1.76 \ \mu m$ \\ \hline
		\textbf{This study} & C57/BL6 & APP/PS1 & AD & 50-64 & \textit{in vivo }2PEF & $5.71 \pm 1.77 \ \mu m$ \\ \hline
		Boero et al.~\cite{boero1999increased} & BALB/C & - & WT & 11 & \makecell[cl]{postmortem \\ optical imaging} & $2.48-2.70 \ \mu m$ \\ \hline
		Drew et al.~\cite{drew2011fluctuating} & C57/BL6 & - & WT & - & \textit{in vivo} 2PEF & $2.9 \pm 0.5 \ \mu m$ \\ \hline
		Blinder et al.~\cite{blinder2013cortical} & C57/BL6 & - & WT & - & \makecell[cl]{\textit{in vivo} optical img., \\postmortem 2PEF} & $2 - 5.3 \ \mu m$ \\ \hline
		Hall et al.~\cite{hall2014capillary} & C57/BL6J & NG2-DsRed & WT & - & \textit{in vivo} 2PEF & $4.4 \pm 0.1 \ \mu m$ \\ \hline
		\makecell[cl]{Gutierrez-Jim{\'a}nez \\ et al.~\cite{gutierrez2016effect}} & C57/BL6 & NTac & WT & 13-15 & \textit{in vivo }2PEF & $4.1-4.5 \ \mu m$ \\ \hline
		Cudmore et al.~\cite{cudmore2016cerebral} & C57/BL6 & \makecell[cl]{Tie2-Cre \\ :mTmG} & WT & \makecell[cl]{13-21, \\ 64, 97} & \textit{in vivo} 2PEF & $5.03 \pm 1.18 \ \mu m$ \\ \hline
		Meyer et al.~\cite{meyer2008altered} & C57/BL6 & APP23 \& -  & AD \& WT & 12-108 & \makecell[cl]{postmortem \\ histology} & $4 - 6 \ \mu m$ \\ \hline
		Tsai et al.~\cite{tsai2009correlations} & Swiss & - & WT & - & \textit{in vivo} 2PEF & $3.97-4.11 \ \mu m$ \\ \hline
		Tsai et al.~\cite{tsai2009correlations} & C57/BL6 & - & WT & - & \textit{in vivo} 2PEF & $3.97-4.11 \ \mu m$ \\ \hline
		Heinzer et al.~\cite{heinzer2004hierarchical} & C57/BL6 & APP23 & WT & 52 & MRA & $14 \pm 5 \ \mu m$ \\ \hline 
		Heinzer et al.~\cite{heinzer2004hierarchical} & C57/BL6 & APP23 & AD & 52 & MRA & $14 \pm 5 \ \mu m$ \\ \hline 
		Heinzer et al.~\cite{heinzer2006hierarchical} & C57/BL6 & APP23 & AD & 44 & $SR\mu CT$ & $8.9 \ \mu m$ \\ \hline
		Heinzer et al.~\cite{heinzer2008novel} & C57/BL6 & - & WT & 16 & $SR\mu CT$ & $5.6 \pm 27.9 \ \mu m$ \\ \hline
		Heinzer et al.~\cite{heinzer2008novel} & C57/BL6 & \makecell[cl]{C3H/He:NSE \\ -VEGF$_{1651}$} & other & 16 & $SR\mu CT$ & $5.5 \pm 29.3 \ \mu m$ \\ \hline
		Serduc et al.~\cite{serduc2006vivo} & Swiss nude & - & WT & 5 & \textit{in vivo} 2PEF & $4 - 6 \ \mu m$ \\ \hline
		V{\'e}rant et al.~\cite{verant2007direct} & Swiss nude & - & WT & 5 & \textit{in vivo} 2PEF & $8.2 \pm 1.4 \ \mu m$ \\ \hline
	\end{tabular}
	\label{tab.diameter_in_literature}
	\end{adjustwidth}
\end{table}

\section*{Conclusions} 
Here, we presented \textit{DeepVess}, a 3D CNN segmentation method together with essential pre- and post-processing steps, to fully automate the vascular segmentation of 3D \textit{in vivo} MPM images of murine brain vasculature. \textit{DeepVess} promises to expedite biomedical research on the differences in angioarchitecture and the impact of such differences by removing the laborious, time consuming, and subjective manual segmentation task from the analysis pipelines in addition to elimination of subjective image analysis results. We hope the availability of our open source code and reported results will facilitate and motivate the adoption of this method by researchers and practitioners.

\section*{Data availability statement}
All data underlying these findings is publicly available at Cornell's eCommons online archive: https://doi.org/10.7298/X4FJ2F1D

\section*{Declarations of interest}
none 


\section*{Supplementary materials}
\renewcommand{\thetable}{S\arabic{table}}%
\setcounter{table}{0}
\renewcommand{\thefigure}{S\arabic{figure}}%
\setcounter{figure}{0}
\renewcommand{\thesection}{S\arabic{section}}%
\setcounter{section}{0}

\normalsize

\paragraph*{Manual 3D segmentation protocol using ImageJ} 
First, we created a new \textit{hyper-stack} (File menu$\rightarrow$New) with 3D voxel size and bit depth similar to the original image (e.g. a 16-bit $1024\times1024\times500$ voxel hyper-stack). The original image and the new hyper-stack were then merged (Image menu $\rightarrow$ Color) into a multi-channel hyper-stack, which contained both the raw data and the segmentation results. 
On each image (in the x-y plane) the expert drew segmentation boundaries using the \textit{free hand} tool and fill function (\textit{F key}) while the second channel is selected using scrollbar. The \textit{Color Picker} and \textit{Channels Tool} (Image menu $\rightarrow$ Color) in addition to the Reverse CZT option (Edit menu $\rightarrow$ Options $\rightarrow$ Miscellaneous) were used to expedite the segmentation process. 

\begin{figure}[!h]
	\centering
	\includegraphics[width=0.75\linewidth]{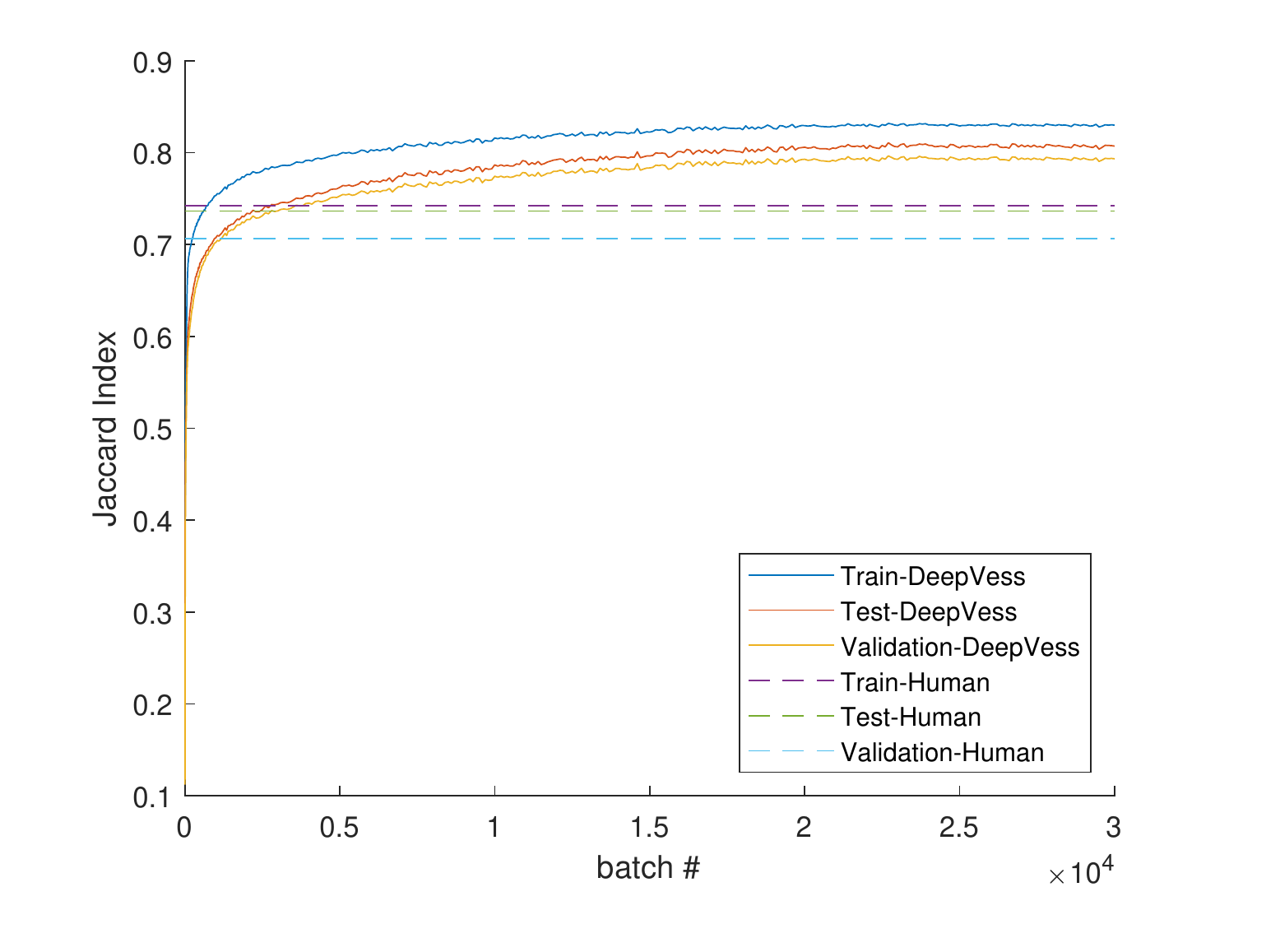}
	\caption{Jaccard as a measure of the model accuracy. The \textit{DeepVess} results surpass the trained human annotator result at all three train, validation, and test datasets. The human annotator and \textit{DeepVess} results are shown in dashed and solid lines respectively. The constant difference between \textit{DeepVess} and the human annotator's results confirm the avoidance of overfitting.}
	\label{figS1.JI_vs_batch}
\end{figure}

\begin{table}[!h]
	\begin{adjustwidth}{-2.25in}{0in}
	\caption{The results of investigating different field of view sizes.}
	\begin{center}
		\begin{tabular}{|l|l|c|}
			\hline
			\textbf{\textit{}} & \textbf{\textit{Architecture }} & \textbf{\textit{FOV}} \\ \hline
			N1 & C 7x7x5 - P - C 5x5 - P - NN & 33x33x5 \\ \hline
			N2 & C 7x7x9 - P - C 5x5 - P - NN & 33x33x9 \\ \hline
			N3 & C 7x7x15 - P - C 5x5 - P - NN & 33x33x15 \\ \hline
			N4 & C 7x7x31 - P - C 5x5 - P - NN & 33x33x31 \\ \hline
			N5 & C 7x7x5 - P - C 5x5 - P - NN & 85x85x5 \\ \hline
			N6 & C 7x7x7 - P - C 5x5 - P - NN & 25x25x7 \\ \hline
			N7 & C 7x7x7 - P - C 5x5 - P - NN & 33x33x7 \\ \hline
			N8 & C 7x7x7 - P - C 5x5 - P - NN & 41x41x7 \\ \hline
			N9 & C 9x9x9 - P - C 5x5 - P - NN & 41x41x9 \\ \hline
		\end{tabular}
	\end{center}
	\begin{center}
		\begin{tabular}{|l|c|c|c|c|c|}
			\hline
			\textbf{\textit{}} &\textbf{\textit{ Sensitivity}} & \textbf{\textit{ Specificity}} & \textbf{\textit{Dice}} & \textbf{\textit{Jaccard}} & \textbf{\textit{MHD}} \\ \hline
			N1 & 93.10\% & 98.15\% & 87.11\% & 77.17\% & 1.38 \\ \hline
			N2 & 87.39\% & 98.87\% & 87.40\% & 77.62\% & 1.15 \\ \hline
			N3 & 91.69\% & 98.31\% & 87.09\% & 77.13\% & 1.61 \\ \hline
			N4 & 89.94\% & 98.21\% & 85.69\% & 74.96\% & 2.19 \\ \hline
			N5 & 91.15\% & 98.23\% & 86.43\% & 76.11\% & 1.46 \\ \hline
			N6 & 90.22\% & 98.61\% & 87.71\% & 78.11\% & 1.03 \\ \hline
			N7 & 91.57\% & 98.49\% & 87.89\% & 78.40\% & 1.20 \\ \hline
			N8 & 91.01\% & 98.34\% & 86.86\% & 76.77\% & 1.85 \\ \hline
			N9 & 93.23\% & 97.61\% & 84.81\% & 73.63\% & 2.38 \\ \hline
		\end{tabular}
	\end{center}
	\label{tab.Results-FOV}
	\end{adjustwidth}
\end{table}

\begin{table}[!h]
	\begin{adjustwidth}{-2.25in}{0in}
	\caption{The results of investigating different architectures.}
	\begin{center}
		\begin{tabular}{|l|l|c|}
			\hline
			\textbf{\textit{}} & \textbf{\textit{Architecture }} & \textbf{\textit{FOV}} \\ \hline
			N10 & C 7x7x7 - P - C 5x5 - P - 2*NN & 33x33x7 \\ \hline
			N11 & 3*C 3x3x3 - P - 3*C 3x3 - P - NN & 33x33x7 \\ \hline
			N12 & 4*C 5x5x5 - P - 3*C 5x5 - P - NN & 41x41x9 \\ \hline
			N13 & 4*C 3x3x3 - P - 3*C 3x3 - P - NN & 41x41x9 \\ \hline
			N14 & C 7x7x7 - P - C 5x5x5 - P - NN & 25x25x25 \\ \hline
			N15 & 3*C 3x3x3 - P - 2*C 3x3x3 - P - NN & 33x33x33 \\ \hline
			N16 & 3*C 3x3x3 - P - 2*C 3x3 - P - NN & 41x41x41 \\ \hline
			N17 & 3*C 3x3x3 - P - 2*C 3x3 - P - NN & 31x31x31 \\ \hline
			N18 & 3*C 3x3x3 - P - 2*C 3x3 - P - NN & 49x49x49 \\ \hline
			N19 & 3*C 3x3x3 - P - 2*C 3x3 - P - NN & 33x33x7 \\ \hline
			N20 & previous architecture for ROI 5x5 & 33x33x7 \\ \hline
			N20P & previous architecture+post proc. & 33x33x7 \\ \hline 
		\end{tabular}
	\end{center}
	\begin{center}
		\begin{tabular}{|l|c|c|c|c|c|}
			\hline
			\textbf{\textit{}} &\textbf{\textit{ Sensitivity}} & \textbf{\textit{ Specificity}} & \textbf{\textit{Dice}} & \textbf{\textit{Jaccard}} & \textbf{\textit{MHD}} \\ \hline
			N10 & 89.61\% & 98.33\% & 86.06\% & 75.53\% & 1.63 \\ \hline
			N11 & 93.71\% & 97.83\% & 86.00\% & 75.44\% & 1.87 \\ \hline
			N12 & 83.78\% & 98.68\% & 84.43\% & 73.05\% & 1.82 \\ \hline
			N13 & 93.45\% & 98.15\% & 87.30\% & 77.46\% & 1.48 \\ \hline
			N14 & 91.57\% & 98.49\% & 87.89\% & 78.40\% & 1.20 \\ \hline
			N15 & 90.29\% & 98.40\% & 86.77\% & 76.63\% & 5.98 \\ \hline
			N16 &  6.31\% & 93.76\% & 7.17\% & 3.72\% & 9.45 \\ \hline
			N17 & 14.82\% & 85.51\% & 10.71\% & 5.66\% & 9.48 \\ \hline
			N18 & 30.40\% & 72.32\% & 13.85\% & 7.44\% & 9.50 \\ \hline
			N19 & 92.89\% & 98.31\% & 87.74\% & 78.15\% & 1.16 \\ \hline
			N20 & 95.15\% & 98.40\% & 89.33\% & 80.71\% & 1.58 \\ \hline
			N20P & 95.09\% & 98.47\% & 89.65\% & 81.24\% & 1.04 \\ \hline
		\end{tabular}
	\end{center}
	\label{tab.Results-Architecture}
	\end{adjustwidth}
\end{table}

\begin{table}[!h]
	\begin{adjustwidth}{-2.25in}{0in}
		\caption{The properties of six 3D images not used for training acquired from different mice included in the second independent dataset.}
		\begin{center}
			\begin{tabular}{|l|c|c|c|c|c|c|}
				\hline
				\textbf{\textit{}} &\textbf{\textit{\makecell[cc]{Image Size \\ (voxel)}}}  & \textbf{\textit{\makecell[cc]{Voxel Size \\ $(\mu m^3)$}}}  & \textbf{\textit{\makecell[cc]{Z evaluation \\ interval}}} & \textbf{\textit{Background}} & \textbf{\textit{Trans gene}} &  \textbf{\textit{Phenotype}}  \\ \hline
				1 & $256\times256\times100$ & $1.14\times1.14\times1$ & 1 $\mu m$ & C57/BL6 & APP/PS1 & AD  \\ \hline 
				2 & $256\times256\times250$ & $0.95\times0.95\times1$ & 25 $\mu m$ & C57/BL6 & APP/PS1 & WT  \\ \hline
				3 & $256\times256\times25$ & $0.95\times0.95\times1$ & 1 $\mu m$ & C57/BL6 & APP/PS1 & WT  \\ \hline
				4 & $256\times256\times25$ & $0.95\times0.95\times1$ & 1 $\mu m$ & C57/BL6 & APP/PS1 & WT  \\ \hline
				5 & $256\times256\times25$ & $0.95\times0.95\times1$ & 1 $\mu m$ & C57/BL6 & APP/PS1 & WT  \\ \hline
				6 & $256\times256\times25$ & $0.95\times0.95\times1$ & 1 $\mu m$ & C57/BL6 & APP/PS1 & WT  \\ \hline
			\end{tabular}
		\end{center}
		\label{tab.SecondImageInfo}
	\end{adjustwidth}
\end{table}

\begin{table}[!h]
	\begin{adjustwidth}{-2.25in}{0in}
	\caption{The results of \textit{DeepVess} and the state-of-the-art methods~\cite{teikari2016deep,cciccek20163d} on the second independent dataset from subjects not used for the model training (Table~\ref{tab.SecondImageInfo}). \textit{DeepVess} surpass both of them in terms of sensitivity, Dice index, Jaccard index, and boundary modified Hausdorff distance (MHD).}
	\begin{center}
		\begin{tabular}{|l|c|c|c|c|c|c|}
			\hline
			\textbf{\textit{}} &\textbf{\textit{ Sensitivity}} & \textbf{\textit{ Specificity}} & \textbf{\textit{Dice}} & \textbf{\textit{Jaccard}} & \textbf{\textit{MHD}} \\ \hline
			Teikari et al.~\cite{teikari2016deep} & 67.7\% & 99.3\% & 74.9\% & 60.6\% & 1.73 \\ \hline 
			3D U-Net \cite{cciccek20163d} & 72.4\% & 99.3\% & 78.5\% & 64.9\% & 1.45 \\ \hline 
			\textit{DeepVess} &\textbf{ 85.5}\% & 98.7\% & \textbf{83.5}\% & \textbf{71.8}\% &\textbf{ 1.41}\\ \hline
		\end{tabular}
	\end{center}
	\label{tab.SecondImageResults}
	\end{adjustwidth}
\end{table}

\begin{figure}[!h]
	\centering
	\includegraphics[width=0.75\linewidth]{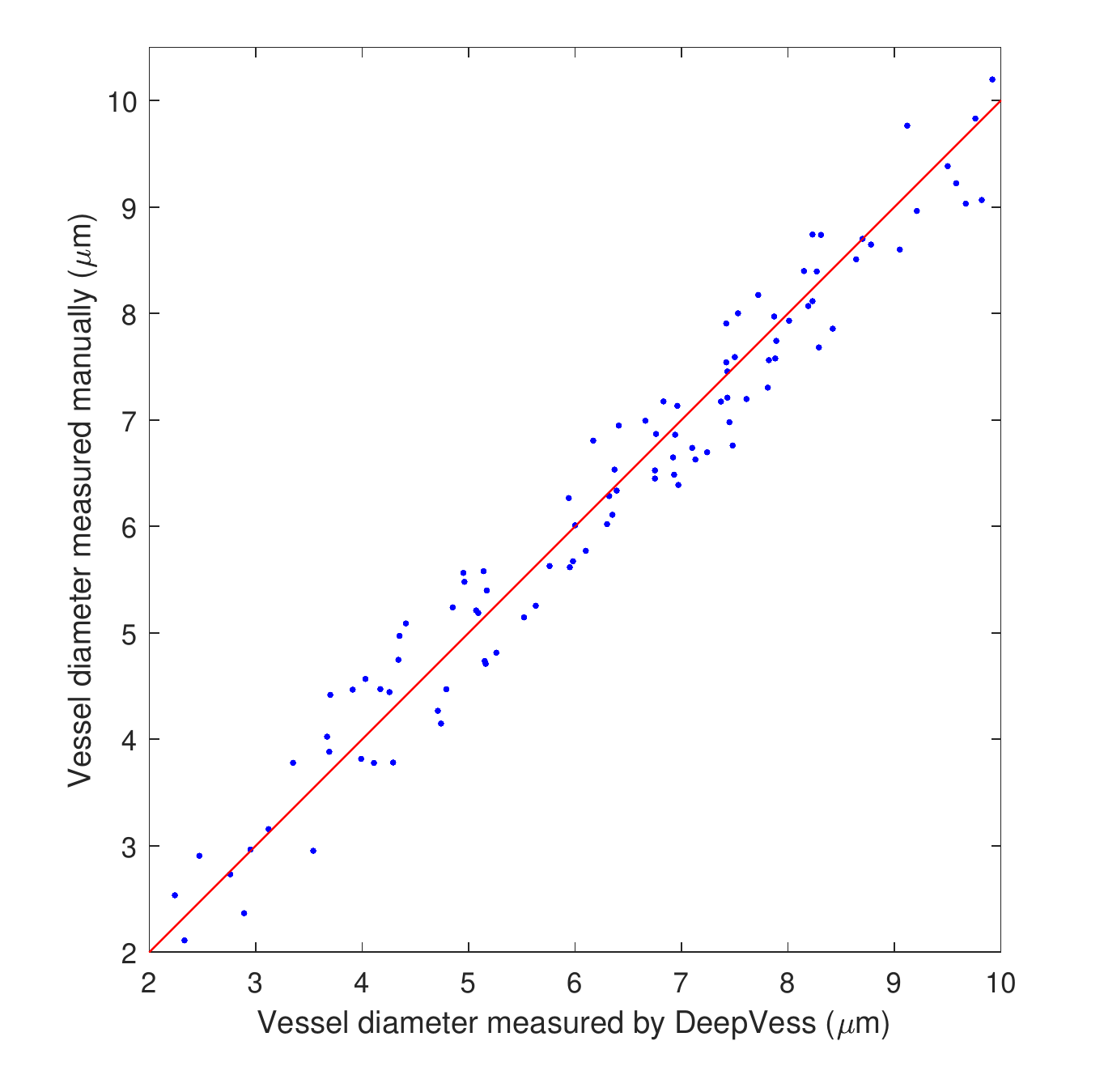}
	\caption{The vessel diameters measured manually in comparison to the \textit{DeepVess's} results. There is no significant difference between two measurements (paired t-test, $n=100$, $p=0.34$).}
	\label{figS2.manual diameter}
\end{figure}

\newpage
\bibliography{DeepVess-bibfile}

\end{document}